\documentclass{article} 
\usepackage{iclr2019_conference,times}
\usepackage{latexsym}
\usepackage{amsmath,amsfonts,amssymb}
\usepackage{bm}
\usepackage{graphicx}
\usepackage{amsfonts}       
\usepackage{nicefrac}       
\usepackage{caption}
\usepackage{svg}
\usepackage{hyperref}

\usepackage{booktabs}
\usepackage{multirow}
\usepackage{url}
\usepackage{subfig}

\usepackage{amsmath,amsfonts,bm}

\def\eqref#1{equation~\ref{#1}}

\def\1{\bm{1}}

\DeclareMathAlphabet{\mathsfit}{\encodingdefault}{\sfdefault}{m}{sl}
\SetMathAlphabet{\mathsfit}{bold}{\encodingdefault}{\sfdefault}{bx}{n}

\newcommand{\sigmoid}{\sigma}

\DeclareMathOperator*{\argmin}{arg\,min}

\usepackage{hyperref}
\usepackage{url}

\newcommand{\ignore}[1]{}

\newcommand{\bfblue}[1]{\textcolor{blue}{\textbf{#1}}}
\newcommand{\WW}{\mathbf{W}}
\newcommand{\XX}{\mathbf{X}}

\newcommand{\parheader}[1]{{\smallskip \noindent \bf #1}}
\newcommand{\balpha}{\bm{\alpha}}

\newcommand{\btheta}{\bm{\theta}}
\newcommand{\bz}{\mathbf{0}}
\newcommand{\bo}{\mathbf{1}}

\newcommand{\bphi}{\bm{\phi}}
\newcommand{\uu}{\mathbf{u}}
\newcommand{\bfv}{\mathbf{v}}

\title{Adaptive Pruning of Neural Language Models for Mobile Devices}

\author{Raphael Tang \& Jimmy Lin \\
David R. Cheriton School of Computer Science\\
University of Waterloo\\
Waterloo, Ontario, Canada \\
\texttt{\{r33tang,jimmylin\}@uwaterloo.ca}
}

\iclrfinalcopy 

\begin{document}
    \maketitle
    \begin{abstract}
Neural language models (NLMs) exist in an accuracy--efficiency
tradeoff space where better perplexity typically comes at the cost of
greater computation complexity. In a software keyboard application on
mobile devices, this translates into higher power consumption and
shorter battery life. This paper represents the first attempt, to our
knowledge, in exploring accuracy--efficiency tradeoffs for NLMs.
Building on quasi-recurrent neural networks (QRNNs), we apply
pruning techniques to provide a ``knob'' to select different 
operating points. In addition, we propose a simple technique to
recover some perplexity using a negligible amount of memory. 
Our empirical evaluations consider both perplexity as well as 
energy consumption on a Raspberry Pi, where we
demonstrate which methods provide the best perplexity--power consumption
operating point. At one operating point, one of the techniques is able to
provide energy savings of 40\% over the state of the art
with only a 17\% relative increase in perplexity.

\ignore{
        On mobile devices, language modeling is widely used for next-word 
        prediction, helping to conserve keystrokes for the user.
        Neural language models currently achieve state-of-the-art results in 
        language modeling; however, their heavy computational requirements can 
        be restrictive, especially when the battery is low. Thus, we describe 
        applying various pruning strategies for reducing the computational 
        footprint of a neural language model at inference-time, without any 
        retraining of the original weights. We 
        propose a technique to train a model where the internal structure is 
        conducive to inference-time pruning. Additionally, we propose a low 
        memory-footprint method to recover further perplexity at pre-defined 
        operating points. We 
        empirically demonstrate that our techniques yield better perplexity 
        ranges than those of baseline approaches. We report power usage figures 
        to support our methods.
}
    \end{abstract}
\section{Introduction}

An emerging application of neural language models (NLMs) is smart
software keyboards on such mobile devices as smartphones and tablets
that provide next-word prediction, allowing users to input entire
words with a single tap. For example,
the apps SwiftKey\footnote{\url{http://www.swiftkey.com/}} and
Swype\footnote{\url{http://www.swype.com/}} both advertise the use of
neural networks for predictions. According to Google Play
Store, SwiftKey has more than 100 million downloads, demonstrating its popularity.

Based on standard metrics such as perplexity, neural techniques
represent an advance in the state of the art in language
modeling~\citep{merity2018analysis}. Better models, however, come at a
cost in computational complexity, which translates to higher power
consumption. In the context of mobile devices, energy
efficiency is, of course, an important optimization objective. A casual web search, for example,
reveals numerous complaints from users of the above apps about battery
drain, indicating that this is not a hypothetical concern.

In reality, neural language models exist in a accuracy--efficiency
tradeoff space. Although this fact has been recognized for
applications such as image recognition~\citep{canziani2016analysis} and keyword
spotting~\citep{tang2017experimental}, to our knowledge no one in the NLP 
community has explored these tradeoffs. All previous papers on NLMs simply report
single-point perplexity figures. In contrast, the high-level goal of our work is to
understand the tradeoffs between neural modeling accuracy and
real-world efficiency constraints:\ in addition to perplexity, NLMs
should be evaluated in terms of FLOPs,\footnote{Convention from literature defines
number of FLOPs as the total number of additions and multiplications.} milliJoule per query (mJ/q), and inference latency. We
conduct exactly such experiments, using the Raspberry Pi (which shares
the same architecture as most mobile devices today) as a more
convenient hardware platform.

Ideally, NLMs should provide a ``knob'' that allows developers to
tune accuracy--efficiency tradeoffs. In
this paper, we explore pruning approaches that take
a pre-trained quasi-recurrent neural network (QRNN; \citealp{bradbury2016quasi}),
representing the state of the art in NLM today, and provides exactly such a knob. Furthermore,
our techniques allow these tradeoffs to be tuned at {\it inference
  time}, which allows a mobile device to adaptively control its
behavior, e.g., favor efficiency at the cost of accuracy when the
battery is low.


Thus, this paper makes the following contributions:\ First, to our
knowledge, we are the first to comprehensively explore accuracy--efficiency tradeoffs
for NLMs with experimental evaluation of energy consumption on
a Raspberry Pi. Second, we evaluate a number of inference-time pruning 
techniques that takes any pre-trained QRNN and provides a tunable 
accuracy--efficiency ``knob''.

    \section{Background and Related Work}

    \subsection{Quasi-recurrent Neural Networks}
        \begin{figure*}[t]
        \centering
        \includegraphics[scale=0.35]{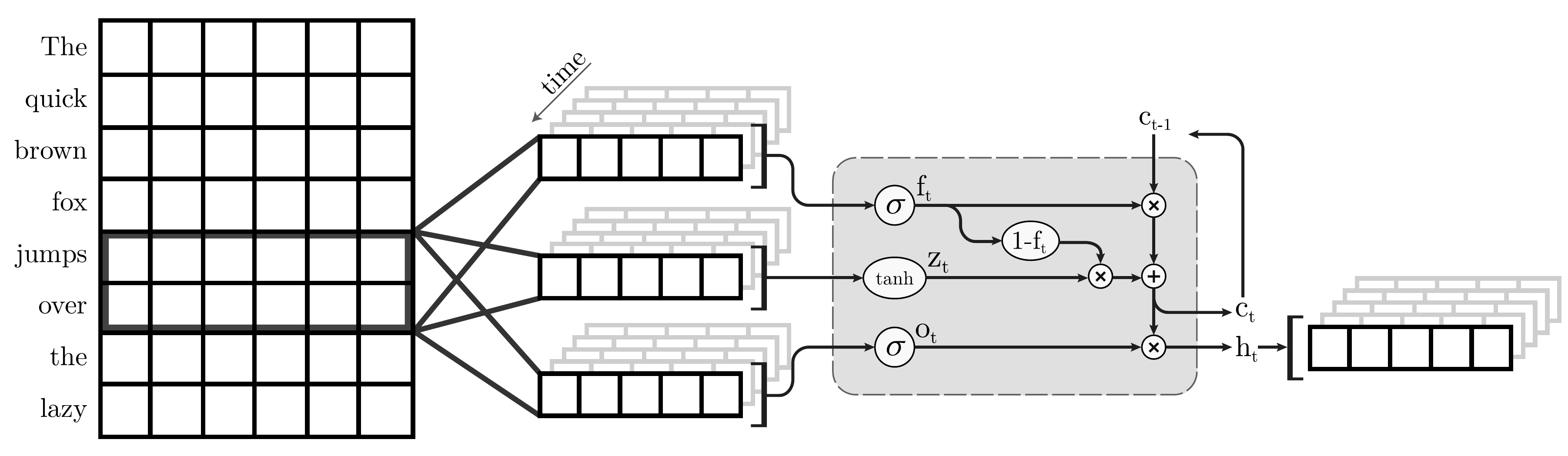}
        \caption{An illustration of the first QRNN layer for language modeling.
        In this visualization, a QRNN layer with a window size of two convolves
        and pools using embeddings from the input. Note the absence
        of recurrent weights.}
        \label{fig:qrnn}
    \end{figure*}
    Quasi-recurrent neural networks (QRNNs; ~\citealp{bradbury2016quasi}) achieve 
    highly 
    competitive perplexity on word-level language modeling datasets, including 
    state-of-the-art perplexity on WikiText-103~\citep{merity2018analysis}. 
    Although applying such techniques as dynamic 
    evaluation~\citep{krause2017dynamic}, Hebbian softmax~\citep{rae2018fast}, 
    and mixture of softmaxes~\citep{yang2017breaking} can produce lower 
    perplexity, our focus is on the recurrent architecture. Thus, we explore 
    the task of pruning QRNNs without using any other extensions.
    
    Each word is encoded as a one-hot vector and then fed into a 
    linear layer, which produces lower-dimensional word embeddings for the QRNN 
    layers. A single QRNN layer consists of two distinct 
    components---convolution and 
    recurrent pooling---that alternate to imitate an LSTM~\citep{lstm}. Given a 
    stacked sequence of inputs $\XX = 
    \bm{x}_1 \oplus 
    \cdots \oplus \bm{x}_n \in \mathbb{R}^{k\times n}$ (e.g., word embeddings 
    in language modeling), the one-dimensional convolution layer is defined as
    \begin{align*}
    \mathbf{Z} &= \text{tanh}(\WW_z \cdot \XX)\\
    \mathbf{F} &= \sigmoid(\WW_f \cdot \XX)\\
    \mathbf{O} &= \sigmoid(\WW_o \cdot \XX)
    \end{align*}
    where $\WW_z$, $\WW_f$, $\WW_o$ are the 
    weights associated with the 
    input, forget, and output gates, respectively,
    \hspace{0.5mm}$\cdot$\hspace{0.5mm} represents 
    a masked convolution along time, and $\sigmoid$ denotes the sigmoid function. 
    For $\WW_{\{z, f, o\}} \in 
    \mathbb{R}^{m\times(k\times r)}$, $m$ is the number of output channels, $k$ 
    is the number of input channels, and $r$ the window size across time. 
    Without loss of generality, we henceforth represent $\WW_{\{z, f, o\}}$ as 
    two-dimensional matrices $\in \mathbb{R}^{m \times s}$, where $s = k \times 
    r$. The 
    outputs are fed into a recurrent pooling layer:
    $$
    \mathbf{c}_t = \mathbf{f}_t \odot \mathbf{c}_{t-1} + (1 - 
    \mathbf{f}_t) \odot \mathbf{z}_t \hspace{10mm}
    \mathbf{h}_t = \mathbf{o}_t \odot \mathbf{c}_t
    $$
    
    \noindent where $\odot$ denotes element-wise product. Altogether, these two 
    layers define a single QRNN 
    layer~(\citealp{bradbury2016quasi}; see Figure \ref{fig:qrnn}). Multiple 
    layers can be stacked for 
    greater expressiveness, where the output $\mathbf{h}_{1:n}$ of the previous 
    layer is the input $\mathbf{X}$ to the current layer.
    
    We tie the weights between the
    input and output layers, as used by \citet{merity2017regularizing} and 
    proposed by \citet{inan2016tying}.
    In addition to improving perplexity, weight tying reduces the number of
    parameters and hence the memory footprint, which is beneficial to our task.
    
    \subsection{Pruning}\label{section:pruning}

    Weight pruning is an effective strategy for reducing the computational 
    footprint of a model. An influential pioneering work, \citet{optimal} 
    proposes to discard weights using a error-approximation approach based on 
    Hessian diagonals. 
    More recent work suggests pruning weights with small 
    magnitudes~\citep{han}, with quantization and Huffman coding as additional 
    steps. However, these approaches introduce irregular sparsity to the 
    weights, and they assume that re-training the weights is feasible. 

In this work, we take a different approach and focus on
techniques that eliminate entire filters. This is because modern
implementations of feedforward evaluation (e.g., im2col and
particularly NEON instruction on ARM processors)\ take advantage of
dense matrix multiplications. Pruning individual weights without
changing the dimensions of the weight matrices has minimal effect on
power consumption---this is confirmed by our initial exploratory
studies on the Raspberry Pi. Hence, we only examine pruning 
techniques that discard entire filters of the convolutional layers:
    
    {\smallskip \noindent \bf Random pruning. }A simple
    baseline~\citep{mittal2018recovering} is random filter
    pruning, where $n\%$ of the filters are randomly pruned, layer-by-layer.
    Interestingly, \citet{mittal2018recovering} find that random pruning
    is competitive with more advanced methods.
    
    {\smallskip \noindent \bf Filter norm. }\citet{channelpruning} propose 
    ranking filters by their $L_1$-norms, and 
    then dropping off $n\%$ of the smallest filters on a layer-by-layer basis.
    \citet{mittal2018recovering} have previously found that $L_1$-norm filter
    pruning~\citep{channelpruning} outperforms a multitude of competing 
    approaches.
    
    {\smallskip \noindent \bf Mean activation norm. } Among other approaches, 
    \citet{errorchannel} suggest pruning filters whose mean activations are 
    small. This approach is especially effective on ReLU, which both creates 
    sparse activations and forces them to be non-negative. 
    
    {\smallskip \noindent \bf $L_0$ regularization. }\citet{l0sparsity} apply 
    $L_0$ regularization to neural networks in order to learn 
    sparse, efficient structures. Formally, define an objective 
    $$\mathcal{R}(\btheta) = \mathcal{L}(\btheta) + 
    \lambda 
    \lVert 
    \btheta 
    \rVert_0 \hspace{10mm}
    \btheta^* = \argmin_{\btheta} \mathcal{R}(\btheta) $$
    where $\mathcal{L}$ is the original loss function and $\btheta$ the 
    weights. The dependence on the hypothesis and training examples has 
    been omitted for brevity. The optimal solution entails a non-differentiable 
    objective and iteration over all $2^{|\btheta|}$ possibilities to choose 
    the best $\btheta^*$; hence,    \citet{l0sparsity} propose the following 
    relaxation of the objective: 
    \begin{gather*}
    \hat{\mathcal{R}}(\btheta, \bphi) = 
    \mathbb{E}_{\mathbf{z} \sim p(\mathbf{z}|\bphi)}\left[\mathcal{L}(\btheta 
    \odot 
    \mathbf{z})\right] + 
    \lambda \sum_{i=1}^{|\btheta|}\left(1 - Q(\mathbf{z}_i \leq 0; \bphi_i)\right)\\
    \btheta^*, \bphi^*=\argmin_{\btheta, \bphi} 
    \hat{\mathcal{R}}(\btheta, \bphi)
    \end{gather*}
    
    \noindent where $\mathbf{z} \sim p(\mathbf{z}|\bphi)$ is a binary 
    discrete random mask parameterized by $\bphi$, and $Q$ is the CDF. 
    Intuitively, for some choice of $\bphi$, the number of active parameters (on average) is 
    penalized. Inspired by the Concrete 
    distribution~\citep{maddison2016concrete}, \citet{l0sparsity} propose the 
    hard concrete distribution for $\mathbf{z}$, further relaxing the discrete 
    random mask into a continuous one:
    \begin{gather*}
    \mathbf{s} = \sigmoid\left((\log \mathbf{u} - \log(1 - \mathbf{u}) + 
    \log\balpha)/\beta\right)\\
    \mathbf{z}=\min(\bo, \max(\bz, (\zeta - \gamma)\mathbf{s} + \gamma))
    \end{gather*}
    
    \noindent where $\mathbf{u} \in \mathbb{R}^{|\btheta|}$ is a continuous 
    random vector such that $\mathbf{u}_i \sim \text{Uniform}\left[0, 
    1\right]$, $\bphi = \log \balpha$ are the mask parameters, and 
    $\gamma = -0.1, \zeta = 1.1, \beta = 2/3$ are scaling hyperparameters. Note 
    that 
    $\beta$ can also be included as part of the mask parameters $\bphi$; we 
    follow \citet{l0sparsity} and fix $\beta = 2/3$. \citet{l0sparsity} then 
    apply the reparameterization trick~\citep{reparam1, reparam2} and make a 
    Monte Carlo approximation to the objective:
    $$\hat{\mathcal{R}}(\btheta, \bphi) = \frac{1}{N} 
    \sum_{i=1}^N\left(\mathcal{L}(\btheta \odot \mathbf{z}^{(i)})\right) + \lambda 
    \sum_{i=1}^{|\btheta|}\left(1 - Q(\mathbf{z}_i \leq 0; \bphi_i)\right)$$
    
    A closed form expression is derived for the penalty, $(1 - Q(\mathbf{z}_i \leq 0; \bphi_i))
    = \sigmoid(\log \balpha_i - \beta \log \frac{-\gamma}{\zeta})$.
    At test time, the following estimator is used: $$\mathbf{z} = \min(\bo, 
    \max(\bz, \sigmoid(\log \balpha)(\zeta - \gamma) + \gamma)$$
    
    \section{Inference-Time Pruning}
    
    In this section, we explain how the various techniques in Section 
    \ref{section:pruning} can be adapted to QRNNs. For the following methods, 
    we assume that a pre-trained model is provided.
    We denote the weights at QRNN layer $l$ 
    as $\WW^{(l)}$. In all methods, we tie the indices across $\WW_z, \WW_f, 
    \WW_o$. For example, if filter $i$ is selected for pruning at layer $l$, 
    then $\WW_{\{z, f, o\}}^{(l)} := \WW_{\{z, f, o\}}^{(l)}[-i, :]$, where 
    $-i$ denotes exclusion of index $i$. This allows the removal of the column 
    $[:, -i]$ in the next layer as well.
    
    \parheader{Random pruning.} We apply random pruning to $\WW_z$, 
    $\WW_f$, and $\WW_o$. That is, we randomly prune filters associated with
    the same indices across the three weights.
    
    \parheader{Filter norm.} We apply filter norm 
    pruning~\citep{channelpruning}, with the filter norms of $\WW_z$ acting 
    as the criteria. We find $\WW_z$ most helpful, since small filter norms 
    should result in small hidden outputs, which is not necessarily the case for 
    $\WW_f$ and $\WW_o$.
    
    \parheader{Mean activation norm.} The hidden output $\mathbf{H} 
    = \mathbf{h}_1 \oplus \cdots 
    \oplus \mathbf{h}_n$ 
    is a natural candidate for collecting mean activation statistics. 
    Intuitively, if 
    $\lVert\mathbf{H}_{:, i}\rVert_1$ is small on average, then the $i^{th}$ 
    filters 
    for $\WW_z, \WW_f, \WW_o$ are less important. Statistics 
    are collected using a single pass of the entire training set. For 
    inference-time pruning, we store the collected statistics.
    
    \parheader{$L_0$ regularization. }Since we are given a pre-trained model 
    and are prohibited from altering the weights, we learn the mask parameters
    only: $\bphi^*=\argmin_{\bphi} \hat{\mathcal{R}}(\btheta, \bphi)$.
    We also enforce the sparsity on entire rows of $\WW_z$, which corresponds 
    to ``group sparsity'' in \citet{l0sparsity}. Specifically, we formulate the 
    regularization on a feature map level instead, with
    $\mathbf{Z}$ as the target:
    $$\mathbf{Z}^{(l)} := \left(\text{diag}(\mathbf{z}^{(l)})\WW_z^{(l)}\right) 
    \cdot \mathbf{X} = \mathbf{Z}^{(l)} \odot \mathbf{z}^{(l)}$$
    
    \noindent $\mathbf{Z}$ is chosen for the 
    property that the $i^\text{th}$ feature map for $\mathbf{h}$ is zero if 
    $\mathbf{Z}_i$ is zero for $\mathbf{c}_0 = \bz$.
    
    This approach entails training and storing extra mask parameters for each 
    operating point.
        However, we find this to be a non-issue for our task, 
    since there are few operating points---three or four at most, out of 
    which we use two for $L_0$ regularization---so the extra storage is 
    negligible.

    \subsection{With Single-Rank Update}
    At specific operating points (e.g., 40\% and 80\% FLOPs), pre-trained 
    weight updates can be  stored and applied at inference-time to recover some 
    perplexity. Suppose $\WW \in 
    \mathbb{R}^{m \times n}$ is a weight matrix in a neural network, and 
    $\WW^* \in \mathbb{R}^{m \times n}$ is some known set of weights that 
    results in a lower loss. Clearly, $\Delta \WW := \WW^* - \WW$ can be stored 
    and added at inference-time to obtain a better neural network. However, it 
    is obvious that this scheme is wasteful, since $\WW^*$ could have directly 
    substituted $\WW$ in the first place.
    
    Sacrificing a negligible amount of storage to recover some perplexity, we 
    propose learning a single-rank weight matrix update $$\Delta\WW 
    := \uu 
    \bfv^\intercal, \uu \in \mathbb{R}^m, \bfv \in \mathbb{R}^n$$
    
    \noindent to each weight in the convolution layers. Specifically, the 
    process is as follows, beginning with a pre-trained model:
    
    \begin{enumerate}
        \item Prune a pre-determined set of filters for some operating point 
        (e.g., 40\% FLOPs).
        \item Initialize the weight updates $\Delta\WW_l = 
        \mathbf{u}^{(l)}{\mathbf{v}^{(l)}}^\intercal, \mathbf{u}^{(l)}_i, 
        \mathbf{v}^{(l)}_i \sim p(\epsilon)$ for each convolution layer $l$, in
        our case $\text{Normal}(0, 0.1)$.
        \item Fixing the existing weights $\WW_l$ for each convolution 
        layer, train a single-rank update such that $\WW^*_l := \WW_l + 
        \Delta\WW_l$, where $\WW^*_l$ is used as the new weight.
        \item Store $\Delta\WW_l$ for use at inference time on the same 
        operating point.
    \end{enumerate}
    
    \section{Experimental Setup}
    
    We evaluate the aforementioned pruning techniques for word-level language 
    modeling on Penn Treebank (PTB)~(\citealp{ptb}; as preprocessed by 
    \citealp{mikolov}) and WikiText-103 (WT103)~\citep{wikitext}. We denote the 
    models for PTB and WT103 as {\tt ptb-qrnn} and {\tt wt103-qrnn}, 
    respectively. 
    
    \subsection{Datasets and Tasks}
    
    For each model, we report word-level perplexity and recall-at-three (R@3), 
    defined as the percentage of top three 
    token--logit outputs that contain the true next token. For example, 
    if \{``cat'', ``dog'', ``baby''\} are the top three predicted tokens for, 
    ``I adopted a \underline{\hspace{6mm}},'' with ``dog'' being the ground 
    truth, then the prediction is correct, regardless of the rank of ``dog''.
    
    {\smallskip \bf \noindent Penn Treebank.} Built from Wall Street Journal 
    articles, Penn Treebank (PTB) is a small yet popular word-level dataset 
    for language modeling. In the standard pre-processed 
    version~\citep{mikolov}, the dataset contains roughly 887K, 70K, and 78K training, 
    validation, and testing tokens, respectively. The number of unique tokens is capped at 
    10,000, yielding a relatively large 4.8\% out-of-vocabulary (OOV) rate.
    
    {\smallskip \bf \noindent WikiText-103.} \citet{wikitext} introduce 
    WikiText-2 and WikiText-103, datasets based on freely available Wikipedia 
    articles. We use only WikiText-103, since WikiText-2 was designed to be 
    similar to Penn Treebank. With 103 million training tokens, 
    WikiText-103 is 103 times as large as PTB. WikiText-103 contains around 
    217K tokens for validation, and 245K for testing. The number of unique 
    tokens is 267K, resulting in a 0.4\% OOV rate, significantly lower than 
    that of PTB.
    
    \subsection{Hyperparameters and Training}   
      
    In all of the models, we chose the hyperparameters as suggested in Merity 
    et al.'s 
    codebase.\footnote{\url{https://github.com/salesforce/awd-lstm-lm}}
    For {\tt ptb-qrnn}, we used a four-layer QRNN with 1550 
    hidden units for each layer and a 400-dimensional embedding. For {\tt 
        wt103-qrnn}, we used a four-layer QRNN with 2500 hidden units and 
    400-dimensional embeddings, along with a tied adaptive 
    softmax~\citep{merity2018analysis}. In both models, the first layer uses a 
    window size of two, while the rest use a windows size of one.
    
    Following \citet{merity2017regularizing}, we also adopted the 
    regularization techniques randomized backpropagation through time, 
    embedding dropout, temporal activation regularization (TAR), activation 
    regularization (AR), and variational dropout. We followed the same training 
    process as well, with non-monotonically triggered ASGD (NT-ASGD) as the 
    optimizer. We use the same hyperparameters as \citet{merity2017regularizing} and
    \citet{merity2018analysis} for each model--dataset pair.
    
    During the training of {\tt wt103-qrnn}, we follow 
    \citet{merity2018analysis}, using a tied adaptive 
    softmax~\citep{grave, merity2018analysis} layer. At inference time, we use 
    a regular softmax instead, since we require R@3.
        
    {\smallskip \bf \noindent Pruning. }We selected a number of distinct 
    operating points that represent discrete points in the accuracy--efficiency 
    tradeoff space. Based on previous work~\citep{tang2017experimental}, floating-point operations (FLOPs) 
    is a good proxy of both 
    energy usage and latency, and so we use FLOPs as a way of selecting our 
    operating points. In $L_0$ regularization, the $\lambda$ decay 
    strength was selected so that the resulting model corresponds to roughly 
    the FLOPs targets: To achieve 80\% and 60\% FLOPs for the model on PTB,
    we used $\lambda = 5.5 \times 10^{-4}, 8.5\times 10^{-4}$, respectively.
    To achieve about 70\% FLOPs on WT103, we chose $\lambda = 6\times 10^{-4}$.

    We trained the hard concrete 
    mask parameters for roughly 5000 steps using Adam with a learning rate of 
    $5\times 10^{-3}$. Since the weight decay penalty is incompatible with the 
    objective, we removed it while training the mask.
    
    For mean activation pruning, which requires some training examples to 
    collect statistics, we used the entire training set for {\tt ptb-qrnn}. 
    Since WikiText-103 is large, we used roughly 10\% of 
    the first training examples for collecting statistics on {\tt wt103-qrnn}.
    
    
    {\smallskip \bf \noindent Single-rank update (SRU). }For the PTB model, the 
    single-rank update was trained for 10 epochs using 
    NT-ASGD~\citep{merity2017regularizing}
    with a non-monotonic interval of three. For WikiText-103, the update was 
    trained for 2000 steps using Adam with a learning rate of $5 \times 
    10^{-3}$. All other hyperparameters were the 
    same as those used during the training stage.
    
    \subsection{Infrastructure Details} 

    We trained all of our models on a commodity machine with a 
    Titan V GPU, i7-4790k CPU, and 16 GB of RAM. We used PyTorch 0.4.0 (commit 
    {\tt 1807bac}) for developing and running our models. We deployed our 
    models on a Raspberry Pi (RPi) 3 Model B (ARM Cortex-A53) running Raspbian 
    Stretch (4.9.41-v7+). Specifically, we copied the trained models over to 
    the RPi, and ran them at the same operating points accordingly.

    We plugged the RPi into a Watts Up Pro meter, a wattmeter that reports 
    power usage at the rate of 1 Hz via a USB cable, which is connected back to 
    the RPi. Evaluating on the test set, we collected power draw statistics on 
    350 next-word predictions, which were averaged to produce a 
    millijoule per query (mJ/q) estimate. We obtained latency estimates in a 
    similar manner by averaging the milliseconds per query (ms/q). Finally, we
    subtracted off the idle power usage of the RPi to obtain a better estimate 
    of the actual power for each query. 

Although our final application is NLMs running on mobile devices such
as smartphones and tablets, there are many challenges to directly
evaluating on such hardware. The Raspberry Pi is a convenient stand-in
since it uses exactly the same ARM processor architecture as nearly
all mobile devices today. Evaluation on the RPi is widely adopted for
research on efficient NNs today~\citep{amato2017deep, tang2017experimental}.

    \section{Results and Discussion}

    In our results for PTB and WT-103, we compare to state-of-the-art results in the past. In general, 
    we find that QRNNs are strong competitors to LSTM approaches, and achieve 
    state-of-the-art perplexity on WikiText-103~\citep{merity2018analysis}.

    \begin{table}[h]
        \centering
        \small
        \begin{tabular}{c l c  c  c  c  c  c  c  c }
            \toprule[1pt]
            \multirow{2}{*}{\raisebox{-3\heavyrulewidth}{\bf \#}} &
            \multirow{2}{*}{\raisebox{-3\heavyrulewidth}{\bf Method}} & 
            \multicolumn{3}{c}{\bf Model Quality} &
            \multicolumn{3}{c}{\bf Footprint} &
            \multicolumn{2}{c}{\bf w/SRU}\\
            \cmidrule(lr){3-5} 
            \cmidrule(lr){6-8}
            \cmidrule(lr){9-10} &
            &  Val. &  Test &  R@3 & 
            \% FLOPs &  ms/q & 
            mJ/q &  Test &  R@3\\
            \midrule
            1 & Skip LSTM & 60.9 & 58.3 & -- & -- & -- & -- & -- & 
            --\\
            2 & AWD-LSTM & 60.0 & 57.3 & -- & -- & 223 & 295 & -- & --\\
            3 & Orig. & 59.0 & 56.8 & 44.7\% & 100\% & 224 & 296 & -- 
            & --\\
            \midrule
            4 & $L_0$ reg. & \bfblue{63.0} & \bfblue{60.7} & \bfblue{43.6\%} & 80\% & 185 
            & 
            227 & 
            \bfblue{59.3} & 
            \bfblue{44.1\%}\\
            5 & $L_0$ reg. & 69.2 & 66.8 & 42.1\% & 60\% & 142 & 
            183 
            & 64.0 & 42.7\% \\
            \midrule
            6 & Random & 68.2 & 66.0 & 42.9\% & 80\% & 182 & 238 & 61.1 & 
            43.8\%\\
            7 & Filter norm & 76.1 & 72.7 & 42.4\% & 80\% & 182 & 238 & 
            66.1 & 43.1\%\\        
            8 & Mean activation & 68.3 & 66.1 & 42.6\% & 80\% & 182 & 238 & 
            61.0 & 43.5\%\\
            \bottomrule[1pt]
        \end{tabular}
        \caption{Select pruning results on Penn Treebank using a 
            4-layer QRNN, along with past results drawn from the original papers. Skip LSTM refers to the four-layer 
            skip LSTM from \citet{melis2017state}, and AWD-LSTM is from \citet{merity2017regularizing}. The four-layer QRNN~\citep{merity2018analysis} is the same model 
            that we use, but we achieve better perplexity following the same methodology. The best
            results of each category are bolded. ``w/SRU'' denotes the results after applying an SRU.}
        \label{table:ptb}
    \end{table}
    
    For PTB, we note that a 20-point increase in perplexity may only correspond to a few 
    points decrease in R@3, showing that perplexity changes 
    on a much different scale than accuracy does (see Table \ref{table:ptb}, 
    rows 3 and 7). Furthermore, lower perplexity does not necessarily imply 
    higher accuracy (see rows 5 and 7), confirming that perplexity alone 
    cannot completely determine the recall. In Table \ref{table:ptb},
    we chose 75 as the cutoff-point for perplexity---further results 
    are illustrated in Figure \ref{fig:wt103ptb}. For WT-103, we observe trends 
    similar to those of PTB; A large drop in perplexity
    corresponds to a much smaller decrease in R@3 (see Table \ref{table:wt103}, rows 3 and 4).

    \begin{table}[t]
        \centering
        \small
        \begin{tabular}{c l  c  c  c  c  c  c  c  c }
            \toprule[1pt]
            \multirow{2}{*}{\raisebox{-3\heavyrulewidth}{\bf \#}} &
            \multirow{2}{*}{\raisebox{-3\heavyrulewidth}{\bf Method}} & 
            \multicolumn{3}{c}{\bf Model Quality} &
            \multicolumn{3}{c}{\bf Footprint} &
            \multicolumn{2}{c}{\bf w/SRU}\\
            \cmidrule(lr){3-5} 
            \cmidrule(lr){6-8}
            \cmidrule(lr){9-10} &
            &  Val. &  Test &  R@3 & 
            \% FLOPs &  sec/q & 
            J/q &  Test &  R@3\\
            \midrule
            1 & Rae-LSTM & 36.0 & 36.4 & -- & -- & -- & -- & -- & -- \\
            2 & 4-layer QRNN & 32.0 & 33.0 & -- & -- & 1.24 & 1.48 & -- & -- \\
            3 & Orig. & 31.9 & 32.8 & 51.5\% & 100\% & 1.24 & 1.48 & -- & 
            -- \\
            \midrule
            4 & $L_0$ reg. & \bfblue{65.8} & \bfblue{65.4} & \bfblue{43.1\%} & 69\% & 0.912 & 
            1.06 & 
            56.9 & 44.7\%\\
            5 & Mean activation & 89.8 & 92.9 & 38.9\% & 70\% & 0.942 & 
            1.10 
            & 
            55.7
            &  46.0\%\\
            6 & Filter norm & 85.9 & 88.2 & 41.7\% & 70\% & 0.942 & 1.10 
            & 
            59.2 & 
            45.4\%\\
            7 & Random & 80.9 & 81.4 & 42.9\% & 70\% & 0.942 & 1.10 & 
            \bfblue{54.2} & 
            \bfblue{46.1\%}\\
            \bottomrule[1pt]
        \end{tabular}
        \caption{Select pruning results on WikiText-103 using a 
            4-layer QRNN, along with past results, drawn directly from the original papers.
            Note that \citet{rae2018fast} primarily explore Hebbian 
            softmax; Rae-LSTM refers to their LSTM model without any extensions. Bolded are the best results
            for each category.}
        \label{table:wt103}
    \end{table}

    \subsection{Accuracy--Efficiency Tradeoffs}
\begin{figure}[h]
        \centering
        \subfloat{
          \centering
          \includegraphics[width=0.5\linewidth]{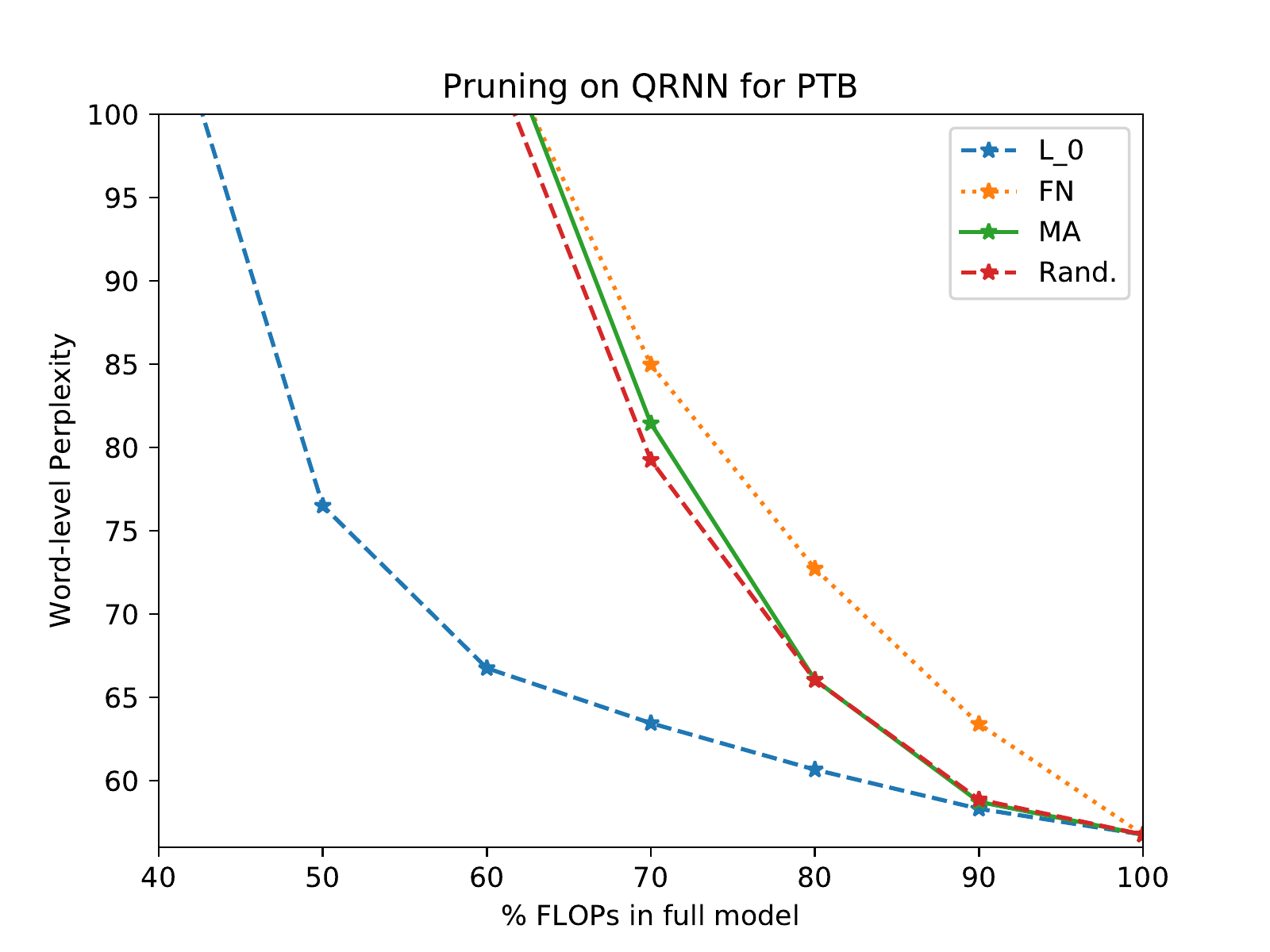}
        }
        \subfloat{%
          \centering
          \includegraphics[width=0.5\linewidth]{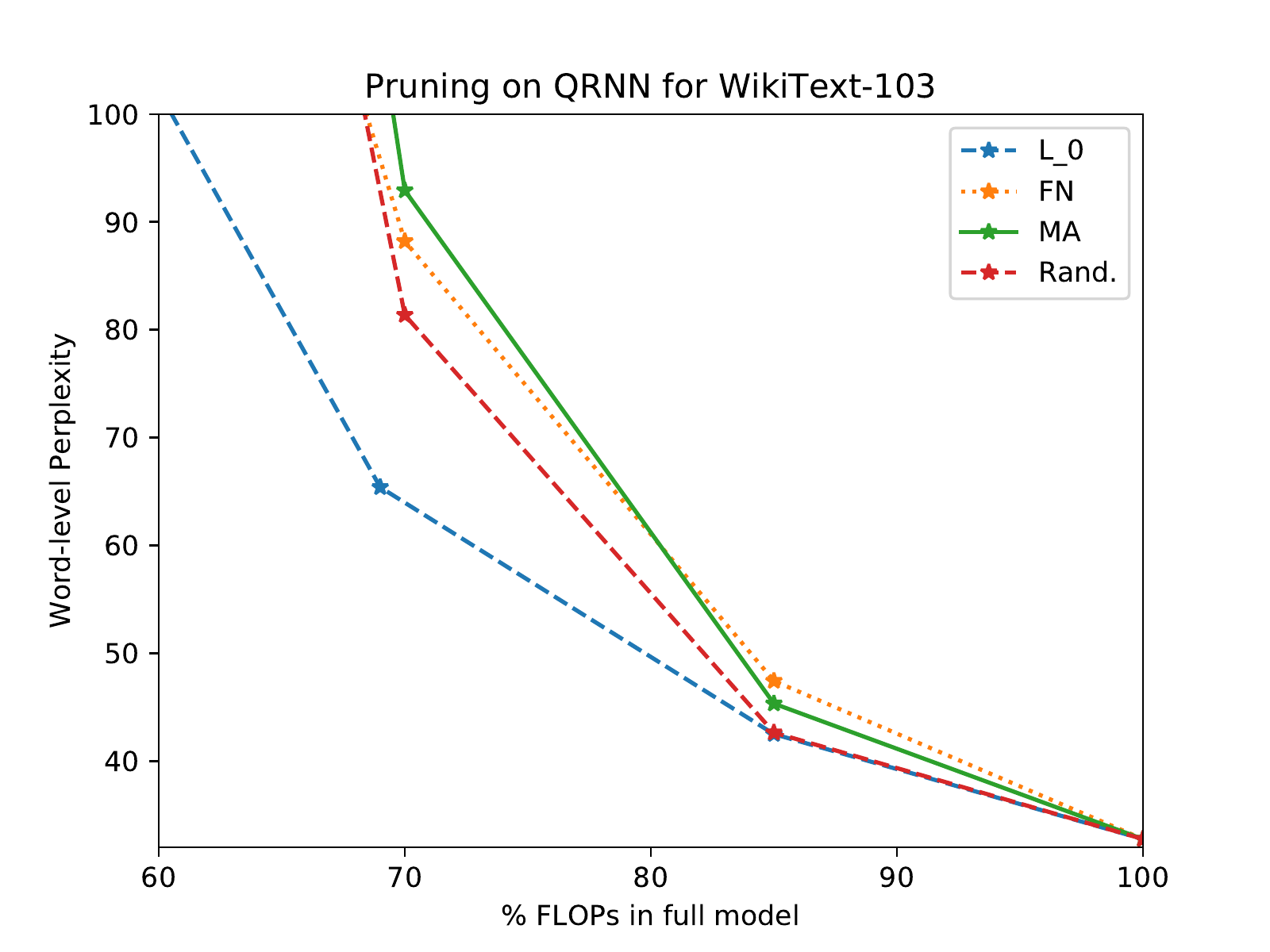}
        }
        \caption{Full experimental results on Penn Treebank and WikiText-103. We illustrate
        the \mbox{perplexity--efficiency} tradeoff space on the test set obtained before applying
        the single-rank update.}
        \label{fig:wt103ptb}
    \end{figure}

    We illustrate the accuracy--efficiency tradeoff space of the PTB and WT103 
    models in Figure \ref{fig:wt103ptb}. For each 
    model, we tabulate the results at fixed intervals according to the 
    approximated percentage of FLOPs, relative to that of the unpruned model. 
    We omit results that exceed 100 in test perplexity, since they are 
    insufficient for language modeling in practice. 

    Surprisingly, random filter pruning is a 
    strong baseline, which supports the findings from 
    \citet{mittal2018recovering}. Random pruning not only outperforms filter 
    norm and mean activation pruning, but also regains perplexity more easily 
    with a single-rank update. From Table \ref{table:ptb} (rows 6--8) and Table 
    \ref{table:wt103} (rows 5--7), we see that random pruning displays 
    equivalent or superior performance to filter norm and mean activation 
    pruning. Interestingly, random pruning achieves the 
    lowest perplexity with a single-rank update (Table \ref{table:wt103}, rows 
    4--7), out of all the baseline approaches on WT103. 
    
    On the other hand, filter norm pruning is relatively weak, doing 
    worse than random pruning in all cases---with or without a single-rank 
    update---suggesting that filter norm pruning has no practical benefit over 
    random pruning. $L_0$ 
    regularization~\citep{l0sparsity} works best, as shown in rows 4--5 in Table 
    \ref{table:ptb} and row 4 in Table \ref{table:wt103}.
    
    
    
    
    In general, testing on Penn Treebank and WikiText-103---two very different 
    datasets---gives us consistent results, thus demonstrating the robustness 
    of $L_0$ regularization~\citep{l0sparsity}, compared to the other pruning approaches.
    \vspace{-3mm}
        \begin{figure}[h]
        \centering\vspace{-6mm}
        \subfloat{
          \centering
          \includegraphics[width=0.185\linewidth]{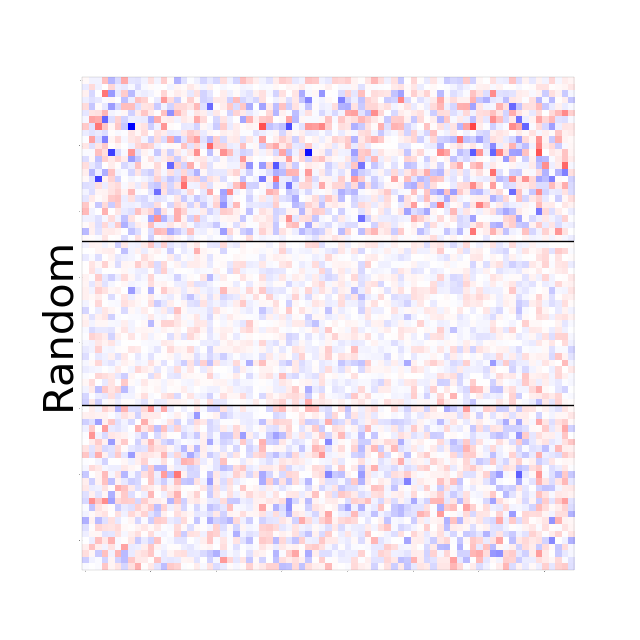}
        }\hspace{-3mm}
        \subfloat{
          \centering
          \includegraphics[width=0.185\linewidth]{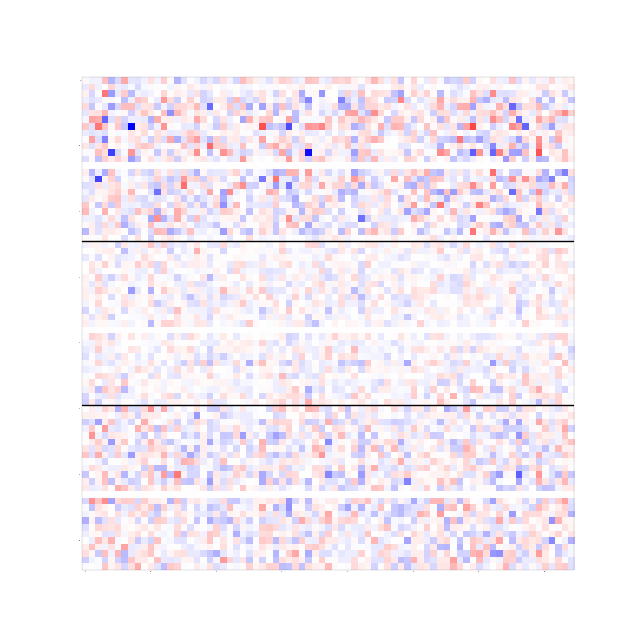}
        }\hspace{-3mm}
        \subfloat{
          \centering
          \includegraphics[width=0.185\linewidth]{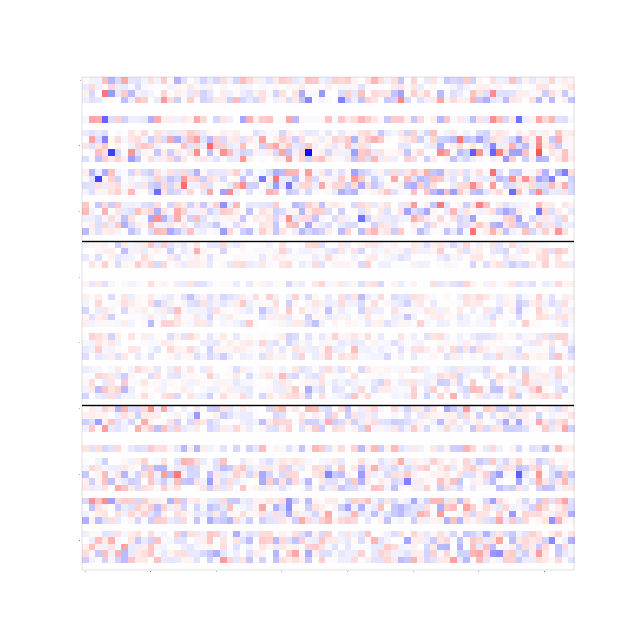}
        }\hspace{-3mm}
        \subfloat{
          \centering
          \includegraphics[width=0.185\linewidth]{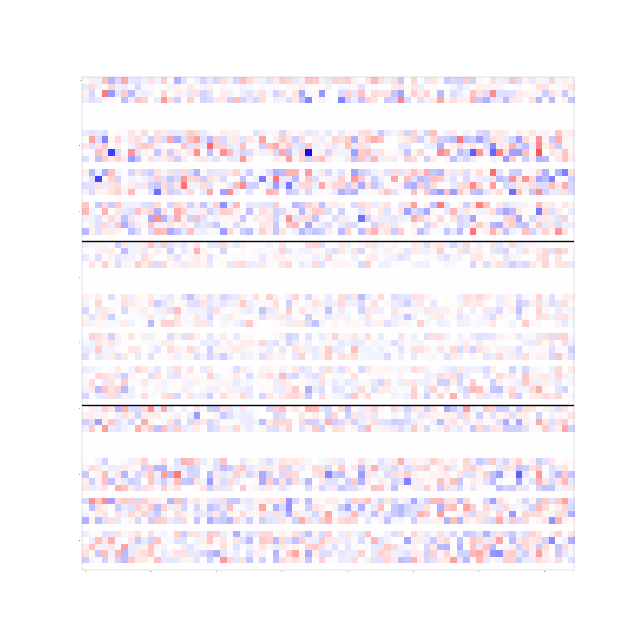}
        }\hspace{-3mm}
        \subfloat{
          \centering
          \includegraphics[width=0.185\linewidth]{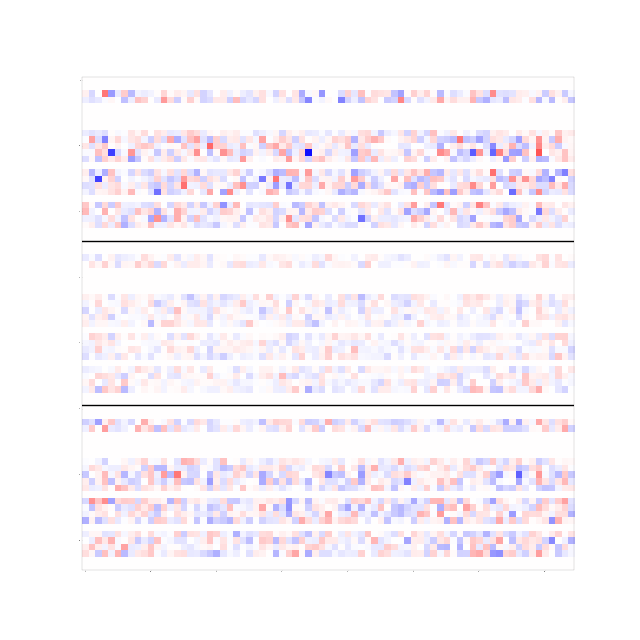}
        }\vspace{-6mm}\hspace{-3mm}
        \subfloat{
          \centering
          \includegraphics[width=0.185\linewidth]{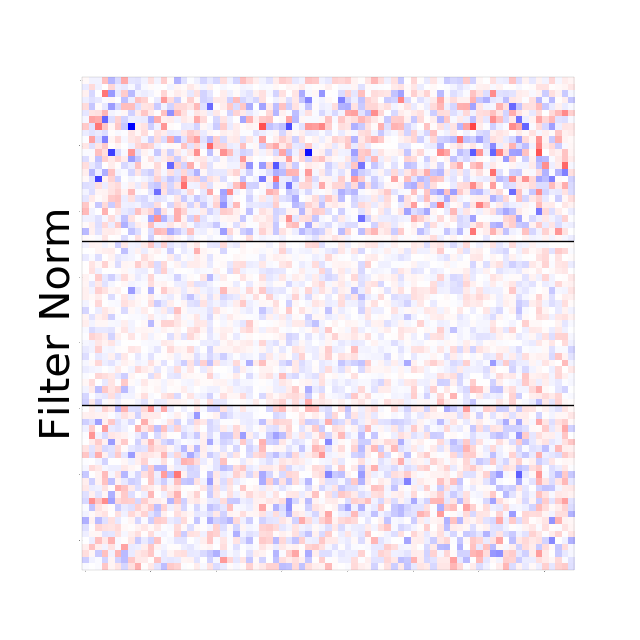}
        }\hspace{-3mm}
        \subfloat{
          \centering
          \includegraphics[width=0.185\linewidth]{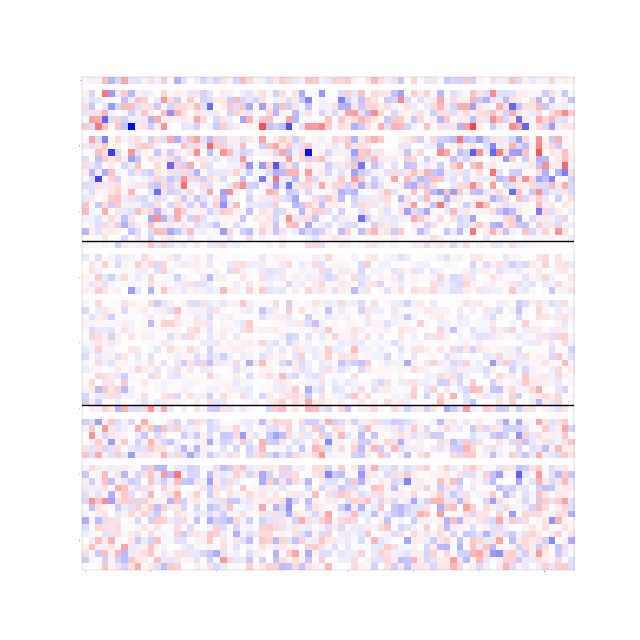}
        }\hspace{-3mm}
        \subfloat{
          \centering
          \includegraphics[width=0.185\linewidth]{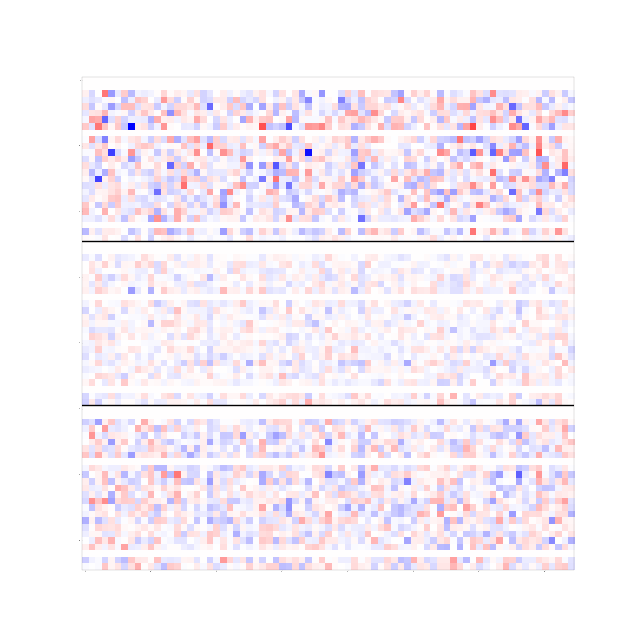}
        }\hspace{-3mm}
        \subfloat{
          \centering
          \includegraphics[width=0.185\linewidth]{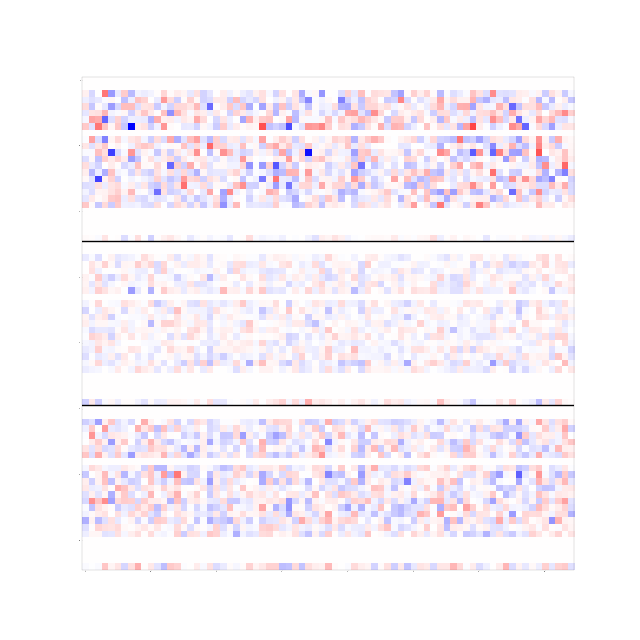}
        }\hspace{-3mm}
        \subfloat{
          \centering
          \includegraphics[width=0.185\linewidth]{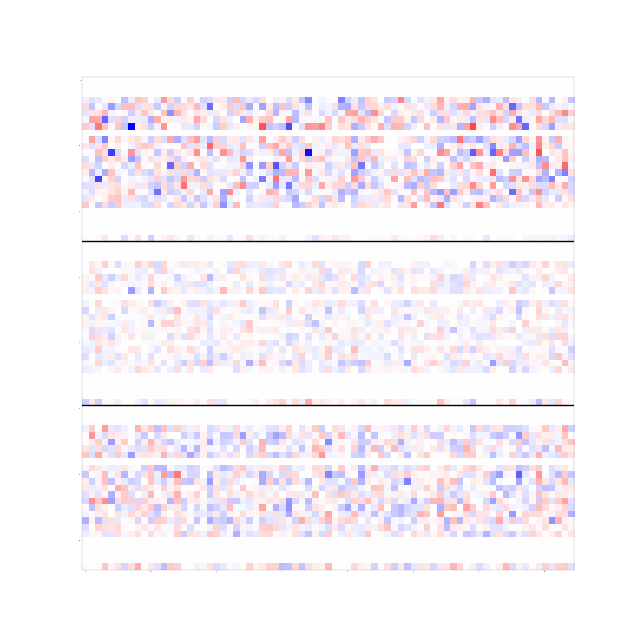}
        }\vspace{-6mm}\hspace{-3mm}
        \subfloat{
          \centering
          \includegraphics[width=0.185\linewidth]{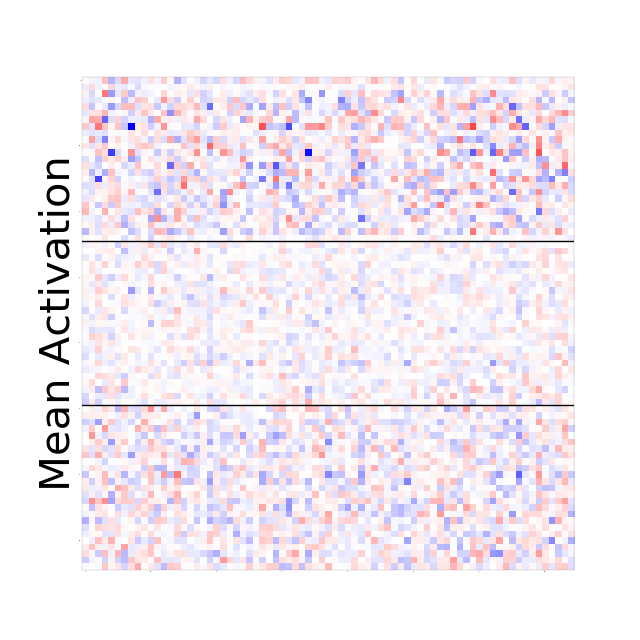}
        }\hspace{-3mm}
        \subfloat{
          \centering
          \includegraphics[width=0.185\linewidth]{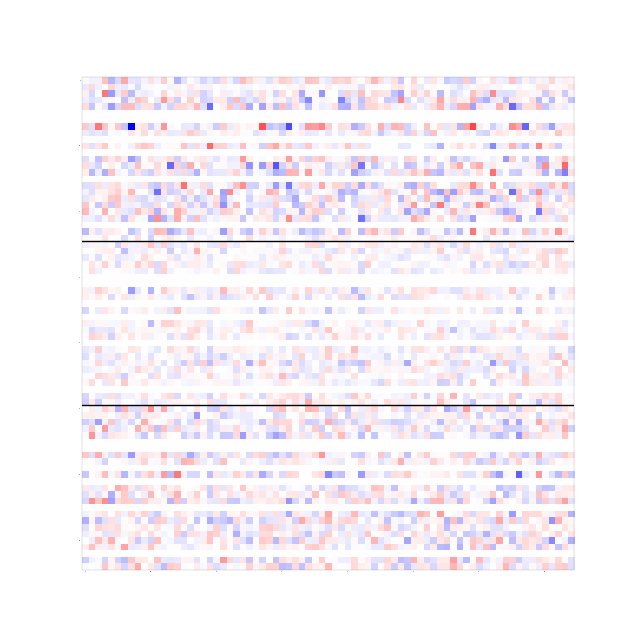}
        }\hspace{-3mm}
        \subfloat{
          \centering
          \includegraphics[width=0.185\linewidth]{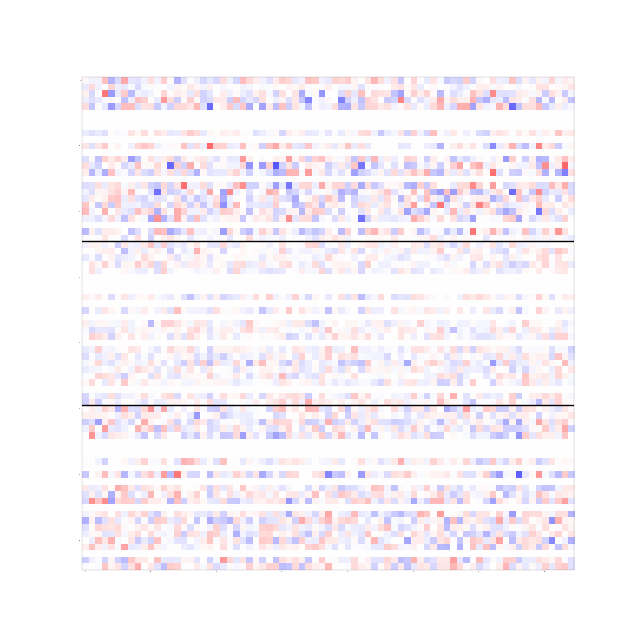}
        }\hspace{-3mm}
        \subfloat{
          \centering
          \includegraphics[width=0.185\linewidth]{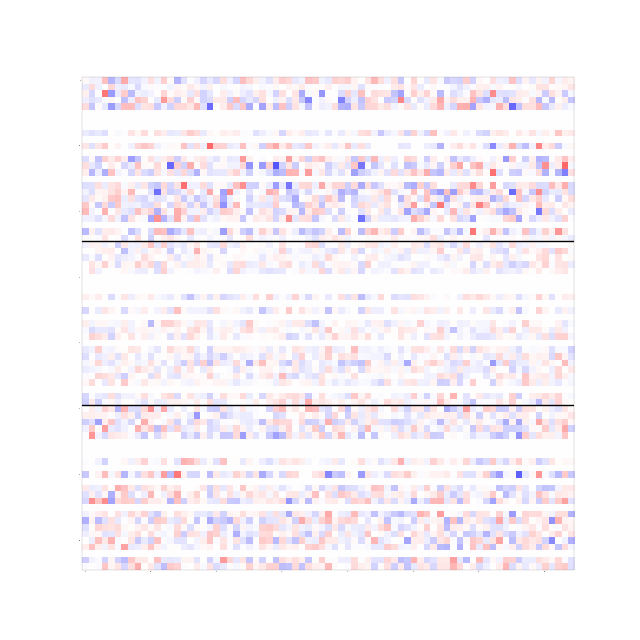}
        }\hspace{-3mm}
        \subfloat{
          \centering
          \includegraphics[width=0.185\linewidth]{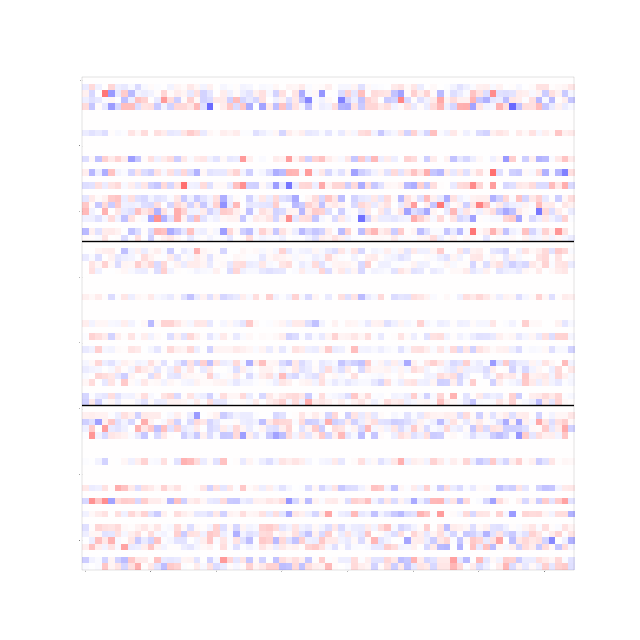}
        }\vspace{-6mm}\hspace{-3mm}
        \subfloat{
          \centering
          \includegraphics[width=0.185\linewidth]{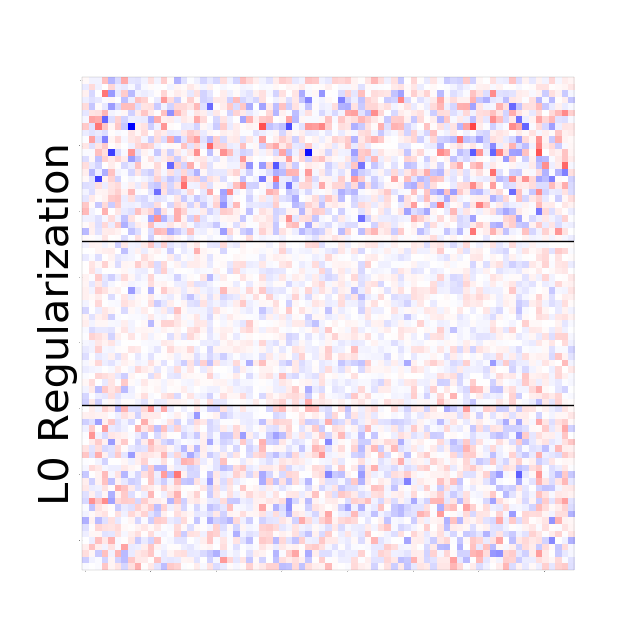}
        }\hspace{-3mm}
        \subfloat{
          \centering
          \includegraphics[width=0.185\linewidth]{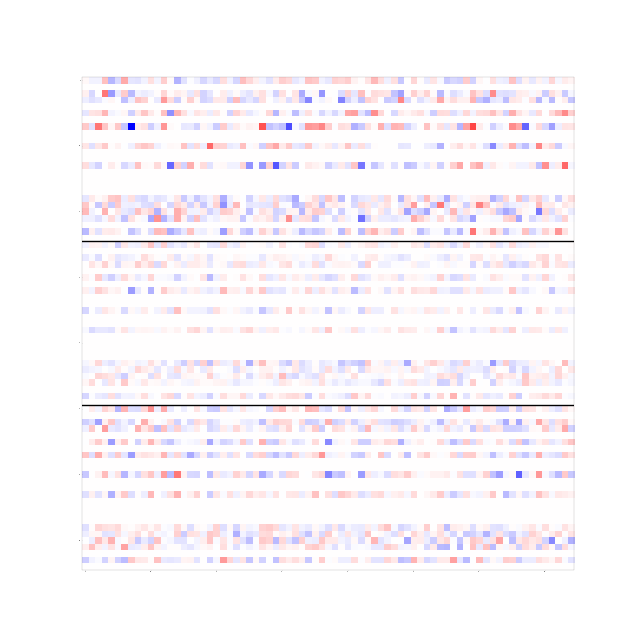}
        }\hspace{-3mm}
        \subfloat{
          \centering
          \includegraphics[width=0.185\linewidth]{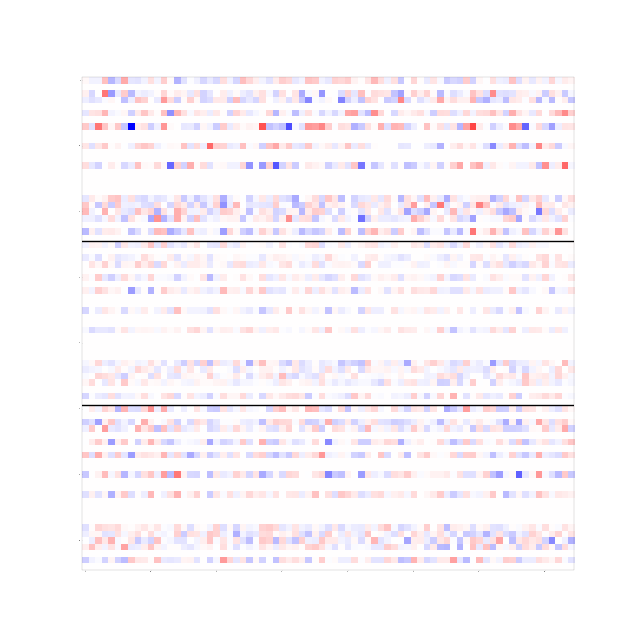}
        }\hspace{-3mm}
        \subfloat{
          \centering
          \includegraphics[width=0.185\linewidth]{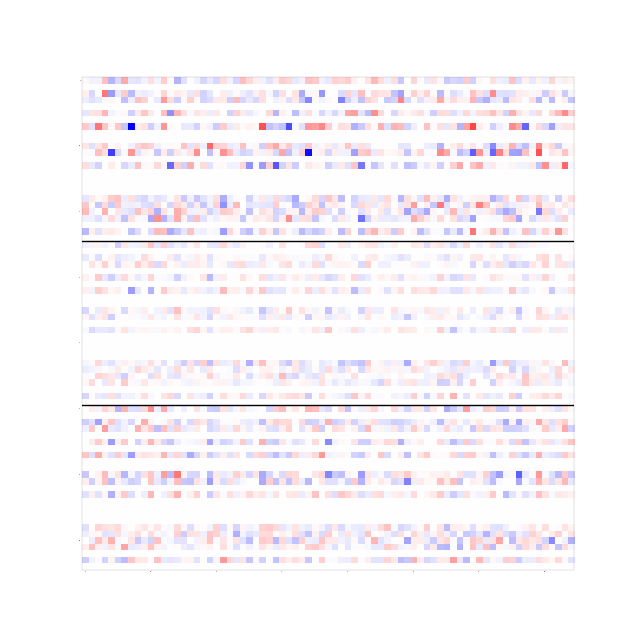}
        }\hspace{-3mm}
        \subfloat{
          \centering
          \includegraphics[width=0.185\linewidth]{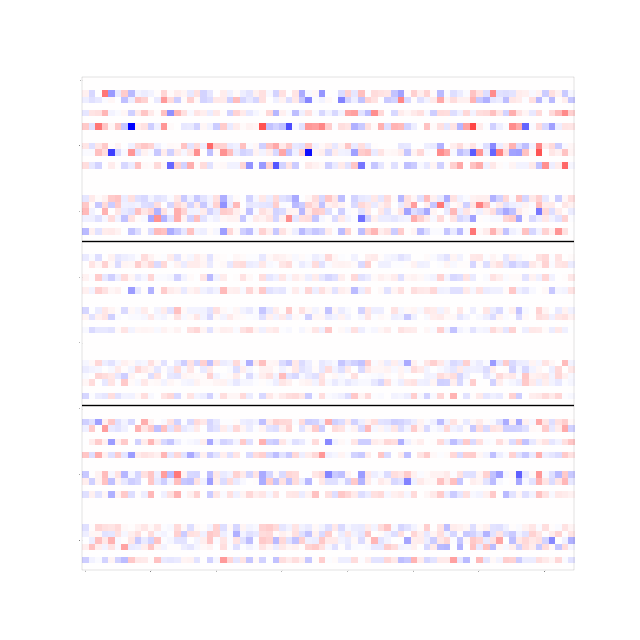}
        }
        \caption{Illustration depicting pruning on a truncated subset of the first layer's weights from the PTB model, where each row
        corresponds to a different technique, and each column a different operating point. From left to right, the operating points are 100\%, 80\%, 70\%, 60\%, and 50\% FLOPs.
        For each of the subfigures, we concatenate from top to bottom the first 25 filters of $\WW_{\{z, f, o\}}$, and
        from left to right the first 75 elements in each filter, yielding square visualizations. All the pruning techniques appear
        to be dropping weights differently---we note that, for $L_0$ regularization (row 4), the dropped weights remain largely constant throughout.}
        \label{fig:visualization}
    \end{figure}
    
    \subsection{Power Usage and Latency}
    
    On the Raspberry Pi, the PTB models are relatively fast, while the 
    WT103 models are high latency, taking over one second (Table 
    \ref{table:wt103}, rows 2--3 and 8) for the full models. For type-ahead 
    prediction on a mobile device, the WT103 models are unsuitable as-is; 
    further steps (e.g., more pruning then re-training, vocabulary reduction, 
    quantization) would be required to deploy the models for practical use. 
    Supporting the findings from \citet{tang2017experimental}, the number of 
    FLOPs scales linearly with latency and power: Full experimental results 
    from Figure \ref{fig:wt103ptb} yield Pearson's $r^2=0.98$ for both 
    latency-- and power--FLOPs measurements, suggesting a strong linear 
    relationship between the number of FLOPs and both latency and power.

    In terms of extra parameters, a single-rank update costs less than 74 KB 
    for {\tt ptb-qrnn}, and less than 120 KB for {\tt wt103-qrnn}. Mean 
    activation statistics requires 20 KB for {\tt ptb-qrnn}, and 30 KB for {\tt 
    wt103-qrnn}. Mask parameters for $L_0$ regularization cost 
    about 20 KB on each power level for {\tt ptb-qrnn}, and 
    30 KB for {\tt wt103-qrnn}. Filter norm pruning and random pruning
    do not require any extra storage.
    
    \section{Conclusion}
    Motivated by the mass adoption of smart software keyboards on mobile 
    devices, we explore the task of inference-time pruning on QRNNs, state-of-the-art 
    neural language models. Starting with existing training-time pruning methods, we 
    extend their usability to QRNNs at run-time, obtaining multiple operating
    points in the accuracy--efficiency tradeoff space. To recover some
    perplexity using a negligible amount of memory, we propose to
    train and store single-rank weight updates at desired operating points.

    \section*{Acknowledgments}
    We are grateful for Meng Dong's work on power measurements and debugging for the RPi
    experiments, and we thank the reviewers for their time and feedback.

\bibliography{adaptive-pruning-main}
\bibliographystyle{iclr2019_conference}

\end{document}